\documentclass[review]{elsarticle}

\usepackage{lineno,hyperref}
\usepackage{tabularx}
\makeatletter
\g@addto@macro{\UrlBreaks}{\UrlOrds}
\makeatother
\modulolinenumbers[1]

\journal{Journal of \LaTeX\ Templates}

\begin{document}

\begin{frontmatter}

\title{Combat COVID-19 Infodemic Using Explainable Natural Language Processing Models}

%% Group authors per affiliation:
\author{Jackie Ayoub}
\address{Industrial and Manufacturing Systems Engineering, University of Michigan-Dearborn \\
4901 Evergreen Road, Dearborn, MI 48128}

\author{X. Jessie Yang}
\address{Industrial and Operations Engineering, University of Michigan \\
1205 Beal Avenue, Ann Arbor, MI 48015}
% \fntext[myfootnote]{Since 1880.}

\author{Feng Zhou\corref{mycorrespondingauthor}}
\address{Industrial and Manufacturing Systems Engineering, University of Michigan-Dearborn \\
4901 Evergreen Road, Dearborn, MI 48128}

\author{Manuscript accepted by Information Processing and Management}
%% or include affiliations in footnotes:
% \author[mymainaddress]{Industrial and Operations Engineering, University of Michigan}
% % \ead[url]{www.elsevier.com}
% % \author[mysecondaryaddress]{Industrial and Manufacturing Systems Engineering, University of Michigan-Dearborn\corref{mycorrespondingauthor}}
\cortext[mycorrespondingauthor]{Corresponding author}
\ead{fezhou@umich.edu}

%\address[mymainaddress]{1205 Beal Avenue, Ann Arbor, MI 48015}
%\address[mysecondaryaddress]{4901 Evergreen Road, Dearborn, MI 48128}

\begin{abstract}
Misinformation of COVID-19 is prevalent on social media as the pandemic unfolds, and the associated risks are extremely high. Thus, it is critical to detect and combat such misinformation. Recently, deep learning models using natural language processing techniques, such as BERT (Bidirectional Encoder Representations from Transformers), have achieved great successes in detecting misinformation. In this paper, we proposed an explainable natural language processing model based on DistilBERT and SHAP (Shapley Additive exPlanations) to combat misinformation about COVID-19 due to their efficiency and effectiveness. First, we collected a dataset of 984 claims about COVID-19 with fact checking. By augmenting the data using back-translation, we doubled the sample size of the dataset and the DistilBERT model was able to obtain good performance (accuracy: 0.972; areas under the curve: 0.993) in detecting misinformation about COVID-19. Our model was also tested on a larger dataset for AAAI2021 - COVID-19 Fake News Detection Shared Task and obtained good performance (accuracy: 0.938; areas under the curve: 0.985). The performance on both datasets was better than traditional machine learning models. Second, in order to boost public trust in model prediction, we employed SHAP  to improve model explainability, which was further evaluated using a between-subjects experiment with three conditions, i.e., text (T), text+SHAP explanation (TSE), and text+SHAP explanation+source and evidence (TSESE). The participants were significantly more likely to trust and share information related to COVID-19 in the TSE and TSESE conditions than in the T condition. Our results provided good implications in detecting misinformation about COVID-19 and improving public trust. 

\end{abstract}

\begin{keyword}
COVID-19, misinformation detection, trust, BERT, DistilBERT, SHAP

\end{keyword}

\end{frontmatter}

%\linenumbers
\section{Introduction}

\emph{“The best way to prevent COVID-19 is actually traditional Chinese medicine”}; \emph{“COVID-19 came from Chinese people eating bat soup”}; \emph{“Coronavirus is an engineered bioweapon”}; \emph{“Coronavirus is just like the flu”} \cite{cdc2020}. Such false and misleading information about COVID-19 has been widely disseminated in digital spaces and is even promoted by famous public figures, including celebrities and politicians. The focus has been going beyond prevention and treatment to include its origin and conspiracy theories. The World Health Organization announced a massive “infodemic” that made it difficult for us to find trustworthy sources and reliable advice amid this horrific pandemic \cite{kassam_disinformation_2020}. What is lacking is how to distinguish true information from false claims timely with rapid changes of the COVID-19 pandemic and provide explanations in order to mitigate the risks associated with COVID-19 to improve public trust through digital spaces \cite {beaunoyer2020covid}. Hence, there is an urgent need to develop prediction models to debunk false claims and to provide timely and trustworthy information to the general public. The failure to meet this need represents an important issue amid the COVID-19 pandemic, as without it both misuse and disuse of such misinformation will continue to exist. 

In order to avoid the spread of misinformation, researchers have been focusing on machine learning-based NLP (Natural Language Processing) techniques. For example, Ozbay and Alatas \cite{ozbay2020fake} used twenty-three supervised machine learning models to identify misinformation. Recently, more successful models based on deep learning techniques have been used to detect misinformation. For example, Aggarwal et al. \cite{aggarwal2020classification} detected misinformation using BERT with very minimal text pre-processing, but obtained very good performance. It was also reported that by April 2020 Facebook removed more than fifty million posts related to COVID-19 since they were classified as misinformation using machine learning-based NLP techniques \cite{sumbaly_using_nodate}. Other big social media companies, including Google and Twitter also removed scammers of ads related to face masks, hand sanitizers, and manipulative posts related to COVID-19 using these deep learning-based models \cite{sumbaly_using_nodate}. Hence, in this study, we proposed an NLP machine learning model based on BERT \cite{devlin2018bert} to detect misinformation about COVID-19. BERT is a deeply bidirectional, unsupervised language model, and was pre-trained using a huge amount of text corpus. One of the disadvantages of BERT is that it is computationally intensive, which might be difficult to be deployed without advanced computational resources. Thus, we made use of DistilBERT which was able to maintain almost similar performance of BERT with fewer parameters \cite{devlin2018bert}.

Despite the efforts using deep learning models in debunking misinformation about COVID-19, there is a lack of research on how to help the general public detect misinformation and improve their trust at the same time. In addition, NLP techniques based on machine learning are usually black-box models. The trust in and acceptance of such models are often compromised without revealing the domain knowledge, i.e., explainability, contained in the data \cite{doshi2017towards}. Compared to other domains, the importance of explainability in decision making with high risks is even greater, such as COVID-19 in the medical area \cite{zhou2020not} and fatigue detection in driving . On the contrary, if the insights captured by such models are revealed, it can help improve trust and acceptance and potentially attain the intended purposes. For example, Gilpin et al. \cite{gilpin2018explaining} showed that explainable machine learning models achieved a higher level of acceptance and trust. Thus, it is crucial for the black-box machine learning models to provide explanations about their decisions. 

In order to improve the trustworthiness of the model, we proposed to explain the reasoning process of distilBERT using SHAP (SHapley Additive exPlanations). SHAP capitalizes on the Shapley value from cooperative game theory \cite{shapley1953value} to calculate individual contributions of the features in the prediction model. It has many desirable properties in explaining machine learning models, including local accuracy, missingness, and consistency \cite{lundberg2017unified}. Furthermore, we designed a between-subjects experiment to evaluate the proposed explainable NLP models in three conditions (i.e., text (T), text+SHAP explanation (TSE), and text+SHAP explanation+source and evidence (TSESE)) in terms of trust and willingness to share the information. As a summary, the contributions of this paper includes 1) building a prediction model based on state-of-the-art NLP techniques, i.e., DistilBERT, 2) explaining model predictions using SHAP in order to improve public trust, and 3) conducting a human-subject experiment to evaluate trust in model prediction and willingness to share information.

\section{Related work}
In order to reduce the negative effects of misinformation, researchers developed different prediction models to automatically detect misinformation. Machine learning models using various features are dominant, such as linguistic features (e.g., styles of writing, subjectivity, and authenticity) and social contextual features (e.g., user characteristics and credibility, content, and social networks) \cite{shu2017fake}. These machine learning-based methods can be further categorized into supervised and semi-supervised methods. Gilda \cite{gilda2017evaluating} evaluated different supervised models on misinformation classification and the best performance was obtained with stochastic gradient descent. Ozbay and Alatas \cite{ozbay2020fake} assessed the performance of twenty-three supervised machine learning models (e.g., logistic model tree, stochastic gradient descent, classification via clustering, bagging, decision tree). The decision tree model achieved the best performance. With the existing computational capabilities and a large amount of data, supervised deep learning models provide better performance compared to traditional machine learning models. Recurrent neural networks (RNNs) and convolutional neural networks (CNNs) were explored by multiple researchers to handle misinformation detection \cite{wang2017liar}. Ma et al. \cite{ma2016detecting} proposed a RNN model with well-designed recurrent units and extra hidden layers to learn the latent features of the contextual information of microblogs over time and their model performed better than online rumor debunking services. Ruchansky et al. \cite{ruchansky2017csi} proposed a hybrid deep RNN model that incorporated three modules, including capture, score, and integrate, which obtained better results in fake news detection. Yang et al. \cite{yang2018ti} proposed a CNN-based model by including text and images as features for the model. Bahad et al. \cite{bahad2019fake} proposed a bi-directional long short-term memory (LSTM) model to detect misinformation. The model was capable of detecting complex patterns in text data by examining a sentence bi-directionally and thus performed better than unidirectional LSTM models. Ma et al. \cite{ma2019detect} proposed a model based on generative adversarial learning to detect misinformation on Twitter, in which a generator was used to produce conflicting information in order for the discriminator to learn better representations to detect rumors. 
For a more detailed review on fake news detection, please refer to \cite{perez2018automatic,zhou2020survey}.

The main limitation of the supervised method is to label a huge amount of  data to train the model, which is laborious and time-consuming. Therefore, in search for an alternative to the supervised method, semi-supervised methods learn domain knowledge from a small set of labeled data on top of a pre-trained unsupervised model. Gaucho et al. \cite{guacho2018semi} proposed a semi-supervised content-based approach to detect misinformation, using tensor embedding and label propagation. Benamira et al. \cite{benamira2019semi} introduced a content-based semi-supervised approach by capturing contextual similarities using a graph learning task. Shu et al. \cite{shu2019beyond} focused not only on news content features but also on publisher bias and user engagement to detect misinformation using a semi-supervised linear classifier to guide the misinformation detection process. Dong et al. \cite{dong2020two} implemented a two-paths semi-supervised deep learning approach based on three CNNs, where both the labeled and unlabeled data were used to train the model. One of the powerful deep learning semi-supervised methods in detecting misinformation is BERT \cite{devlin2018bert}. BERT is composed of two stages, i.e., unsupervised pre-training and supervised fine-tuning. Aggarwal et al. \cite{aggarwal2020classification} showed that BERT outperformed LSTM and gradient boosted tree models even with minimal text pre-processing. To improve the performance of BERT, Jwa et al. \cite{jwa2019exbake} proposed a model that classified the data using weighted cross-entropy. They pre-trained BERT on additional news data and obtained better results than BERT.

However, the BERT model is computationally expensive and contains millions of parameters (i.e., 110 million parameters for BERT base and 340 million parameters for BERT large) \cite{devlin2018bert}, which makes it difficult to apply in real time without accelerated hardware, such as GPUs and TPUs. In contrast, in this paper, we attempted to address this problem by distilling the knowledge from BERT to detect misinformation \cite{sanh2019distilbert}. DistilBERT was shown to be 60\% faster than BERT while retaining over 95\% of BERT’s performance \cite{sanh2019distilbert}. Another difficulty is the lack of labeled data in fine-tuning the model for a domain specific task. Therefore, we proposed a data augmentation method using back-translation, which helped to double the size of the training data collected by ourselves and to improve the model performance. For example, Xie et al. \cite{xie2019unsupervised} showed that sentences generated by back-translation reached a significant improvement in text classification. 

Furthermore, in the case of fact checking related to COVID-19 claims, both understanding and trust are necessary for the adoption of the predictions. Very few studies focused on improving trust in model prediction by incorporating model explanation \cite{laihuman2019}. However, according to Rudin et al. \cite{rudin2019stop}, an inaccurate explanation limits the trust in the model. Therefore, it was suggested that we should not only show the model performance but also include explanations about the predictions  \cite{laihuman2019}. 
There are two approaches to explain a machine learning model including example- and feature-based methods \cite{laihuman2019}. The example-based approach is based on criticisms and prototypes \cite{kim2016examples}. This method was proved to improve human understanding and reasoning. For example, Shu et al. \cite{shu2019defend} proposed a sentence-comment co-attention sub-network to detect fake news and used the top-k user comments as contextual information to explain why they were fake. However, example-based methods are often limited to improve the interpretability when the claims do not have contextual information \cite{molnar_chapter_nodate}, where claims about COVID-19 lack specific contexts to tell if they are true or fake. In the feature-based approach (e.g., SHAP), each feature is characterized by an importance value for a particular prediction \cite{lundberg2017unified}. For example, Reis et al. \cite{reis2019explainable} examined a large and diverse set of features of fake news and found some features were very effective to detect certain types of fake news, which were used to explain model decisions to help detect fake news.
The feature-based approach can provide two major advantages, including global and local interpretability. The global interpretation aims to show how much each feature contributed to the overall prediction and the local interpretation explains individual predictions, which tends to be more helpful to improve user understanding and trust. Therefore, we proposed to use SHAP to explain the prediction of DistilBERT locally to improve trust and acceptance as a feature-based explanation method. To further evaluate the explanations provided by SHAP, a between-subjects experiment was designed with three conditions i.e., T, TSE, and TSESE.

\section{Methods}

\begin{figure*}[hbt!]
\centering
\includegraphics[width=1\textwidth]{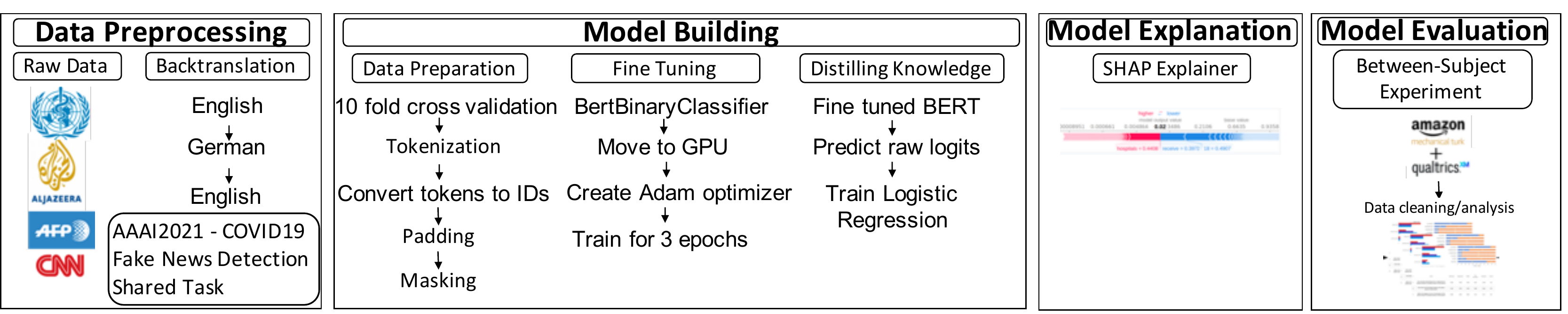}
\caption{Summary of the proposed misinformation detection process.}
\label{Fig. 1}
\end{figure*}

The summary of the proposed method is illustrated in Figure~\ref{Fig. 1} with the following steps:

(1) Data Preprocessing: We collected a dataset and manually labeled the data with fact checking from different trustworthy sources. Then, we doubled the size of the dataset using back-translation. In addition, we included another large dataset for AAAI2021 - COVID-19 Fake News Detection Shared Task \cite{patwa2020fighting} to further test our proposed method.

(2) Model Building: We transformed (i.e., tokenization, padding, masking) the dataset into the shape needed to fine-tune the BERT model. After that, the  PyTorch-Pretrained-BERT library was used to build the BERT model. Then the BERT model was fine-tuned with our labeled data or the COVID-19 Fake News dataset \cite{patwa2020fighting}. Then, we distilled the knowledge from the BERT model by training a logistic regression model.   

(3) Model Explanation: To improve user trust in the distilled BERT model, SHAP was used to explain the predictions locally. 

(4) Model Evaluation: To evaluate user trust in the model predictions and the provided SHAP explanations, we conducted a between-subjects experiment with three conditions on Amazon Mechanical Turk (AMT, Seattle, WA, www.mturk.com). 

\subsection{Data Preprocessing}
We collected and labeled our own misinformation of COVID-19 dataset with lateral reading and verification \cite{wineburg2019lateral} and fact checking from different trustworthy websites.
We collected our claims (in sentences rather than in long news articles, see Table \ref{table: claimExamples}) in English about COVID-19 from well-edited sources, including Cable News Network (i.e., CNN, www.cnn.com), Word Health Organization (i.e., WHO, www.who.int), Centers for Disease Control and Prevention (i.e., CDC, https://www.cdc.gov), and Aljazeera (www.aljazeera.com), and facts check websites, including Snopes (www.snopes.com), FactCheck (www.factcheck.org), and Poynter (www.poynter.org/covid-19-poynter-resources). The reasons that we collected the claims from these sources are that 1) these sources are trustworthy and 2) individual tweets, posts, or reports without verification are less reliable, especially at the early stage of the COVID-19 pandemic when knowledge about the virus was not well-established. 
% Please add the following required packages to your document preamble:

\begin{table}[]
\caption{Examples of collected COVID-19 claims}
\label{table: claimExamples}

\begin{tabularx}{\textwidth}{lllX}
%\begin{tabularx}{\textwidth}{|l|X|}

\hline
Label & Source    & Date           & Claims\\
\hline
Fake & FactCheck &
  Jan. 28, 2020 &
  “Chinese spy team” working in a Canadian government lab sent “pathogens to the Wuhan facility” prior to the coronavirus outbreak in China. \\
Fake  & Snopes    & March 2, 2020  & Sales of Corona beer dropped sharply in early 2020 because consumers mistakenly associated the brand name with the new coronavirus    \\
Fake  & Poynter   & Jan. 22, 2020  & Chinese influencer caused the new coronavirus outbreak after eating bat soup.                                                         \\
Fake  & WHO       & March 20, 2020 & COVID-19 only affects the old.                                                                                                        \\
True  & CNN       & April 9, 2020  & There's no evidence to support the theory that 5G networks cause COVID-19 or contribute to its spread                                 \\
True  & WHO       & March 20, 2020 & Being able to hold your breath for 10 seconds or more without coughing or feeling discomfort DOES NOT mean you are free from COVID-19 \\
True  & CDC       & April 11, 2020 & Stay home for 14 days after your last contact with a person who has COVID-19.                                                         \\
True  & FactCheck & March 16, 2020 & Gargling water with salt won’t ‘eliminate’ coronavirus    \\                             
\hline                                           
\end{tabularx}%

\end{table}

Then, we developed a back-translation augmented method to increase the sample size of our own collected data by using a high-quality translation app (www.deepl.com/en/translator). Back-translation is simply translating a text back to the original language (i.e., English) after translating it into another language (i.e., German) \cite{xie2019unsupervised}. This resulted in new sentences differed from what we started with. For example, using the back-translation technique, two original claims, i.e., “Consuming boiled ginger with an empty stomach can kill the coronavirus” and “Several viral tweets purporting that snorting cocaine would sterilize one's nostrils of the coronavirus spread around Europe and Africa” became “Eating cooked ginger on an empty stomach can kill coronavirus” and “In Europe and Africa, several viral tweets spread claiming that snorting cocaine would rid one's nostrils of coronavirus”, respectively. Although the new claims had nearly the same meaning as the original ones, the key words and some of the word orders were different. We collected 984 claims (575 true and 409 fake) about COVID-19, and doubled the sample size with back-translation. 

The COVID-19 Fake News dataset \cite{patwa2020fighting} had 10,700 claims. However, only the training and validation dataset with 8,560 claims (4,480 true and 4,080 fake) were available and used in this paper. They were directly used in the BERT and DistilBERT models. However, for traditional machine learning models (see Table \ref{table:performance}), we prepared both this dataset and our own labeled dataset using tokenization, lemmatization, removing stop words and punctuation, and converting the textual representation into a vector space model using the term frequency-inverse document frequency. 

\subsection{Model Building}
\textbf{Fine-tuning with BERT:} BERT produced state-of-the-art results in a wide variety of NLP tasks (e.g., question answering, translation, and text classification). It was first pre-trained on a huge amount of text data (800M words from the BooksCorpus and 2,500M words from the English Wikipedia) \cite{devlin2018bert}. The basic transformer relied on an encoder to read the text and a decoder to produce a prediction. To prepare the needed input to the BERT encoder, the data was passed into three embedding layers including a token, segment, and position embedding layers. In the first step of the processing, sentences were tokenized and after that each input token was passed through a token embedding layer to transform it into a vector representation of fixed dimension (i.e., 768-dimensional vector). Additionally, extra classification [CLS] and separator [SEP] tokens were added to the start and end of the tokenized sentence to serve as an input representation and a sentence separator for the classification task. The segment embedding layer helps in classifying a text given a pair of input texts. The positional embedding layer learns the relative position of tokens in a sentence using a sinusoidal function. The final input embedding is a summation of the three embeddings. The summed input was passed to the transformer. 
In this study, we used the PyTorch-Pretrained-BERT library to build the BERT model. Then, we fine-tuned its linear layer and the sigmoid activation to obtain the predictions with the labeled COVID-19 dataset. During the fine-tuning process, Adam optimizer was used with a learning rate of $3\times 10^{-6}$ and a batch size of 12. We fine-tuned the model on the collected COVID-19 dataset for three epochs.

\textbf{DistilBERT:} It is an approximation method of BERT that uses only 60\% of  the number of BERT model parameters (i.e., 66 million parameters instead of 110 million). The main benefit of DistilBERT is its capability of almost reproducing the behavior of BERT by compressing the big BERT model. In this study, we made use of the knowledge distillation process in DistilBERT, defined as a compression technique in which the student (i.e., DistilBERT) is trained to mimic the teacher’s behavior (i.e., BERT) \cite{sanh2019distilbert}. The BERT predictions were first used to train a smaller model, DistilBERT, by learning the inner representation with raw predictions (i.e., predictions before the final activation function) rather than the hard target probabilities. Then, the knowledge was transferred to the student with a cross-entropy on the raw target probabilities of the teacher and the distillation loss of the training process is as follows:
\begin{equation}
L= \sum_it_i*log (s_i)  
\end{equation}
where \(t_i\) and \(s_i\) are the probabilities estimated by the teacher and the student, respectively. We ran a distillation for three epochs with a learning rate $3\times 10^{-6}$ and a batch size of 12 using the Adam optimizer on Google Colaboratory.

\textbf{Logistic regression:} It is a supervised classification technique that is characterized by a logistic function to model a probability (i.e., sigmoid function) of a prediction given a set of features. In this paper, we distill the knowledge from the BERT model by training a logistic regression model, which might result in a slight loss of accuracy in order to improve the explainability of the model predictions using SHAP. 

\subsection{Model Explanation}
SHAP was used in this paper to explain the output of the DistilBERT model by assigning each feature with an importance value related to a particular prediction \cite{lundberg2018consistent}. SHAP is built on Shapley value derived from coalitional game theory \cite{shapley1953value}, in which each player is assigned with payouts depending on their contribution to the total payout when all of them cooperate in a coalition. It combines optimal allocation with local explanations using the classic Shapley values. Studies have shown that it is often easier for the users to trust prediction models not only by providing what the prediction is, but by also providing why and how the prediction is made \cite{kovalerchuk2018toward,ayoub2021modeling}. In this paper, SHAP is used to explain DistilBERT (logistic regression) model. The units of the SHAP values are in the log-odds space, which was then transformed into predicted truth probabilities (see Figure 2). 

The SHAP value for the $i$-th feature-value set is calculated as follows: 
\begin{equation}
\varphi_i= \beta_i . (x_i - E[x_i]) 
\end{equation}
where \(\beta_i\) is the weight corresponding to feature \textit{i}, \(x_i\) is a feature value, \(E[x_i]\) is the mean effect estimate for feature \textit{i}. For example, if we want to predict if a given claim represents a misinformation or not, each feature (i.e., word) will have its contribution to push the final prediction away from the base value (see Figure 2). By aggregating all the features for one instance marginalized over all other features that are not included in the set $S$, we can calculate the overall SHAP value  \cite{lundberg2018consistent},
\begin{equation}
f_\mathbf{x}(S) = E[f(\mathbf{x})|\mathbf{x}_S] = \int \hat f(x_1,...,x_P)dP_{x\notin S}-E(\hat f(\mathbf{x})), 
\end{equation}
where $P$ is the number of the words in the instance, and $S$ is the set of non-zero indexes of words in the dataset and $\mathbf{x} = [x_1,...,x_P]$. $E[f(\mathbf{x})|\mathbf{x}_S]$ indicates the expected value of the function conditioned on the subset $S$ of the input words in the model. 
Then, according to the coalitional game theory \cite{shapley1953value}, the Shapley value of the $i$-th feature-value set is defined as its contribution to the payout, weighted and summed over all possible feature-value combinations as follows:
\begin{equation}
\phi_i(f) =  \sum _{S \subseteq N \backslash \{i\}} \frac{|S|!(P-|S|-1)!}{P!} (f_\mathbf{x}(S \cup \{i\}) - f_\mathbf{x}(S)),\\
\label{phi}
\end{equation}
where $N$ is the set of all the input words and $N \backslash \{i\}$ indicates the set that does not include $i$-th word. In order to estimate both $E[f(\mathbf{x})|\mathbf{x}_S]$ and $\phi_i(f)$ efficiently, we adopted the TreeSHAP algorithm proposed in \cite{lundberg2018consistent}.

\subsection{Model Evaluation}

To evaluate the provided SHAP explanations, we conducted a between-subjects experiment on AMT. AMT is a web-based survey company, operated by Amazon Web Services. The survey was created with Qualtrics (Provo, UT, www.qualtrics.com), web-based survey software, and was integrated with AMT. We investigated participants’ trust in model predictions and their willingness to share the provided information in three conditions, i.e., T, TSE, and TSESE, as shown in Figure 2 using a 7-point Likert scale. The source and evidence information was manually collected during our own the data collection process and we included the source and evidence and information as a third condition, i.e., TSESE, to further test if such information was needed, comparing with the TSE condition. 

\begin{figure}[hbt!]
 \center
\includegraphics[width=1\columnwidth]{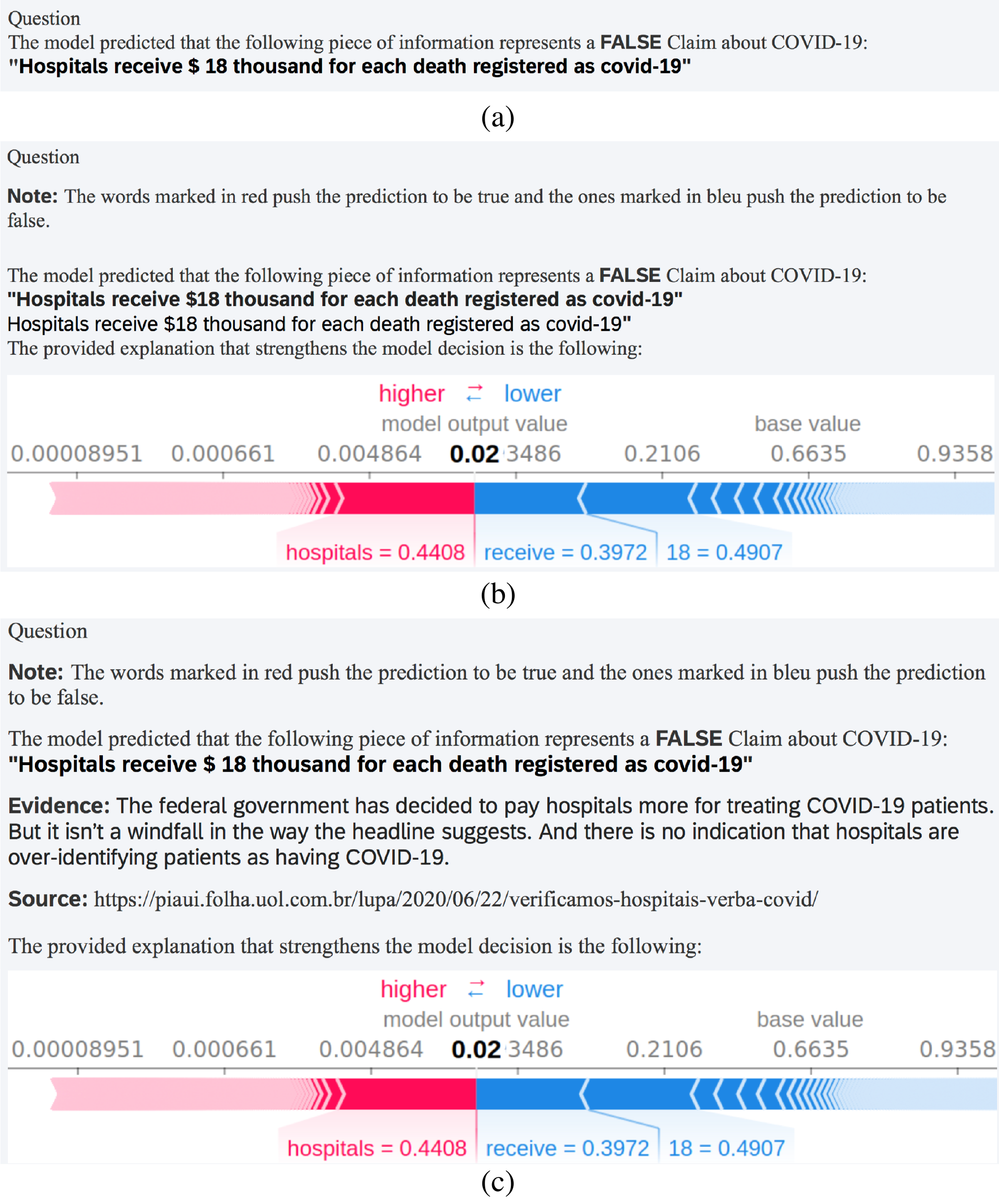}
  \label{fig2}
\caption{Three conditions involved in the between-subjects experiment: (a) Condition T (Text); (b) Condition TSE (Text+SHAP Explanation ); (c) Condition TSESE (Text+SHAP Explanation+Source and Evidence.} 
\end{figure}

\begin{figure*}[hbt!]
 \center
{\includegraphics[width=1\textwidth]{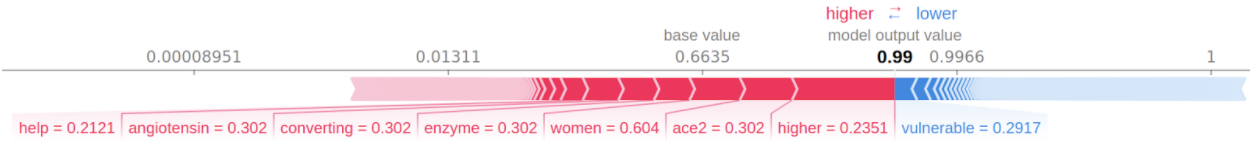}}
\label{Fig. 3(a)}
\center\small{(a)}
{\includegraphics[width=1\textwidth]{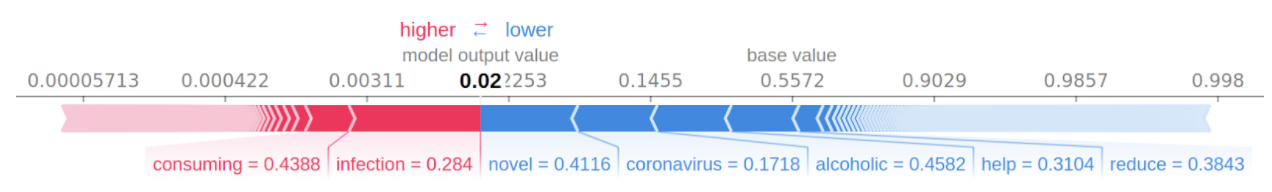}}
  \label{Fig. 3(b)}
\center\small{(b)}
\caption{Examples of a (a) true claim: (\emph{“Men have higher concentrations of angiotensin-converting enzyme 2 (ACE2) in their blood than women, which may help to explain why men are more vulnerable to COVID-19 than women”}) and (b) false claim: \emph{“Consuming alcoholic beverages may help reduce the risk of infection by the novel coronavirus”} explained by SHAP.} 
\end{figure*}

We designed a between-subjects experiment where each participant was randomly assigned to one of the conditions. In condition T, participants were given a classifier prediction (i.e, true, false) about a claim related to COVID-19 (see Figure 2(a)). In condition TSE, in addition to the claim and the model prediction, SHAP explanation was provided to increase the model transparency in making predictions (see Figure 2(b)). In addition to the information provided in condition TSE, evidence and source of the claims were presented to the participants in condition TSESE (see Figure 2(c)). In the three conditions, participants reported their degree of trust in model predictions by answering \emph{“Based on the given explanation, what is your degree of trust in the model prediction”}) and their willingness to share by answering \emph{“Based on the given explanation, how much are you willing to share this claim with your friends and/or families”}) using a 7-point Likert scale. In the given SHAP explanations, participants were explained with three main points, including 1) the output value which represents the predicted truth probability (i.e., if it is close to 0 the claim is more likely to be false, whereas if it is close to 1 it is more likely to be true), 2) a base value which represents the mean predicted truth probability, and 3) words represented in red (i.e., pushing the prediction to be true) and blue (i.e., pushing the prediction to be false). In each condition, there were 10 claims, including 5 true and 5 false, about COVID-19 randomly selected from our own labeled dataset. After going through a training session and correctly answering two attention-check questions, the participant was eligible to take part in the survey. One qualitative question was also designed  at the end of each condition in the survey, which asked the participants to state the reasons behind trusting/distrusting the model predictions and willingness/unwillingness to share the information. A total number of 300 participants in the USA filled in the survey with 100 in each condition. In order to complete the survey, participants needed to be 18 years old and above. We further removed participants who did not answer the third attention question correctly at the end of the survey. We ended up with 84 participants in condition T and 80 participants each in conditions TSE and TSESE. Participants were compensated with \$1 upon completion of the survey. 

\section{Results}
\subsection{Model Performance}
The performance of the different tested models, including precision, recall, F1 score, accuracy, and area under the receiver operating characteristic curve (short for AUC) using a 10-fold cross-validation process is shown in Table \ref{table:performance}. The BERT model had a slightly better performance (see Table \ref{table:performance}) than the DistilBERT model while the DistilBERT model is more efficient and has better explainability to improve its trustworthiness using SHAP (see Figure 3). Since our model distilled BERT with a much simpler model, i.e., logistic regression, which was much more efficient and without losing much performance compared to BERT, we compared the performance of the augmented DistilBERT with other traditional machine learning models that were also more efficient than BERT. We used Python in Google Colaboratory. The augmented DistilBERT method performed the best among all the selected traditional machine learning models, including classification tree, logistic regression, and random forest (see Table \ref{table:performance}). 
\begin{table*}[tb]
\footnotesize
\centering
\caption{Summary of model performance on our own collected dataset}
\begin{tabular} {c c c c c c} 
 \hline
 Model & \begin{tabular}[c]{@{}c@{}}Precision \\ (False/True)\end{tabular} & \begin{tabular}[c]{@{}c@{}}Recall\\ (False/True)\end{tabular} & \begin{tabular}[c]{@{}c@{}} F1-score \\(False/True)\end{tabular} & Accuracy & AUC \\ \hline 

BERT & 0.998/0.990 & 0.985/0.998 & 0.991/0.994 & 0.993 & 0.999 \\ 
DistilBERT (Logistic) & 0.772/0.876 & 0.836/0.824 & 0.803/0.849 & 0.829 & 0.887\\
Aug-BERT & 0.994/0.994 & 0.991/0.996 & 0.993/0.995 & 0.994 & 0.999 \\
\textbf{Aug-DistilBERT
(Logistic)} & 0.961/0.980 & 0.972/0.972 & 0.967/0.976 & 0.972 & 0.993\\
Classification Tree & 0.721/0.794 & 0.707/0.805 & 0.714/0.800 & 0.764 & 0.756\\
Logistic Regression & 0.835/0.894 & 0.853/0.880 & 0.844/0.887 & 0.869 & 0.922\\
Random Forest & 0.775/0.864 & 0.817/0.831 & 0.795/0.848 & 0.825 & 0.920\\
Aug-Classification Tree & 0.872/0.918 & 0.886/0.908 & 0.879/0.913 & 0.899 & 0.897\\
Aug-Logistic Regression & 0.949/0.957 & 0.939/0.964 & 0.944/0.961 & 0.954 & 0.992\\
Aug-Random Forest & 0.930/0.947 & 0.925/0.950 & 0.928/0.949 & 0.940 & 0.987\\  
\hline
BERT* & 0.998/0.989 & 0.988/0.998 & 0.993/0.994 & 0.993 & 1.000 \\ 
\textbf{DistilBERT
(Logistic)*} & 0.923/0.952 & 0.949/0.928 & 0.936/0.940 & 0.938 & 0.985\\
Classification Tree* & 0.897/0.900 & 0.889/0.907 & 0.893/0.903 & 0.898 & 0.900\\
Logistic Regression* & 0.926/0.941 & 0.936/0.931 & 0.931/0.936 & 0.934 & 0.984\\
Random Forest* & 0.880/0.947 & 0.946/0.883 & 0.912/0.914 & 0.913 & 0.977\\
 \hline
\end{tabular} \\
\small{Note “Aug-” denotes the model was augmented by extra data using back-translation tested with our own labeled dataset. Models with "*" indicate the performance on the COVID-19 Fake News Detection dataset \cite{patwa2020fighting}} 
\label{table:performance}
\end{table*}

\subsection{SHAP Explanation}
To show how SHAP explained individual predictions, we randomly selected two observations as shown in Figure 3. The figure shows the different features contributing to pushing the predicted truth probability from the base value. Factors pushing the prediction to be true are shown in red (i.e., words in red increase the predicted truth probability) while those pushing the prediction to be false are shown in blue (i.e., words in blue decrease the predicted truth probability). The first example (see Figure 3(a)) represents a true claim with a predicted truth probability of 0.99. Words that contributed to producing the given prediction are “help”, “angiotensin”, “converting”, “enzyme”, “women”, “ace2”, and “higher”.  The second example (see Figure 3(b)) represents a false claim with a predicted truth probability of 0.02. The words that contributed to producing the given prediction are “novel”, “coronavirus”, “alcoholic”, “help”, and “reduce”. These words help explain why the model predicted such results in order to improve its trustworthiness.

\subsection{Survey Results}
A one-way ANOVA model was used to analyze the survey data with a significance level of $\alpha = 0.05$. 
A post-hoc analysis was used with a Tukey HSD correction. Figure 4 summarizes the mean and standard error of trust and willingness to share the information under the three conditions. The main effects of the three conditions on trust (${F}(2, 241) = 5.628, \textit{p} = .004$) and willingness to share the information ($\textit{F}(2, 241) = 10.730$, $\textit{p} = .000$) were significant. 
Trust in the model decision was shown to be significantly higher in the TSE condition ($\textit{p} = .031$) and the TSESE condition ($\textit{p} = .005$) than the control condition. Willingness to share was shown to be significantly higher in the TSE condition ($\textit{p} = .001$) and the TSESE condition ($\textit{p} = .000$) than the control condition. However, there were no significant differences between the TSE and TSESE conditions both for trust and willingness to share.

\begin{figure}[tb]
\centering
\includegraphics[scale=0.5]{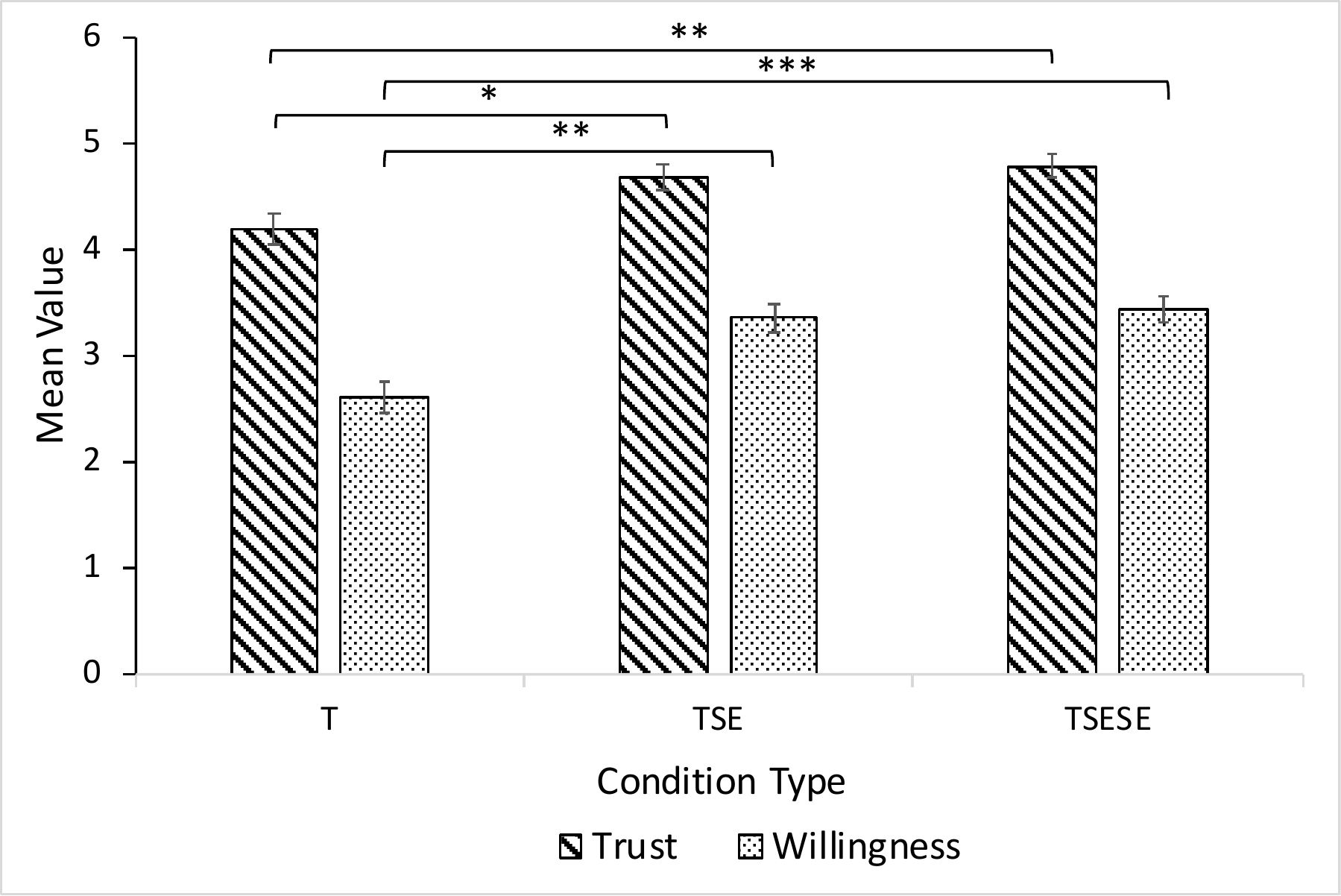}
\caption{The effects of three conditions on trust and willingness to share the information. Note  “T”, “TSE”, and  “TSESE” denote “text”, “text+SHAP explanation”, and “text+SHAP explanation+source and evidence”, respectively; “*”, “**”, and “***” indicate  \textit{p} $<$ .05, \textit{p} $<$ .01, and \textit{p} $<$ .001, respectively. 
}
  \label{Fig. 7}
\end{figure}

\section{Discussions}
\subsection{DistilBERT-based NLP Models }
We showed previously in the results (see Table \ref{table:performance}) that the performance of DistilBERT was reasonably well compared with BERT while having 40\% fewer parameters. In addition to the good performance, DistilBERT was 60\% faster than BERT. 
We also showed that distilling the knowledge from BERT by training a logistic regression model outperformed other traditional machine learning models (e.g., classification tree, logistic regression, and Random Forest). One of the reasons for this performance was that DistilBERT was built on BERT, which learned deep representation of the words by pre-training on contextual representation using a large corpus with bidirectionality, whereas the traditional models used the term  frequency-inverse  document  frequency. A model with good performance is important, especially in the situation of the COVID-19 pandemic, since participant’s trust in model prediction can be improved with higher predicted accuracy \cite{laihuman2019}.

\subsection{SHAP-Explanations and User Trust}
Compared to the control condition (i.e., the T condition), we showed that participants’ trust and willingness to share was significantly enhanced by adding SHAP explanations in the TSE condition. This result proved the effectiveness of SHAP explanation to help improve trust in COVID-19 related claims. This was also supported by participants' qualitative responses, such as \emph{“The predictions confirm my beliefs”}. Furthermore, by adding the source and evidence of the information in the TSESE condition, participants’ trust and willingness to share information were also significantly higher than those in the control condition (one participant stated in the TSESE condition, \emph{“I trusted the predictions if the claims were backed up by reliable resources and evidence”}). Such results were consistent with previous research that providing more explanations (e.g., feature importance, predicted probabilities, sources, and evidence) improved participant’s trust in the model \cite{xie2019unsupervised}.
The increase in willingness to share information is also related to the increase in trust in the information as Mosleh et al. \cite{mosleh2020self} showed that self-reported willingness to share information on social media reflected the actual intentions of trust, which further supported the effectiveness of our proposed model. 

However, there was no significant difference between the TSE and TSESE conditions in terms of trust and willingness to share. In order to investigate the reasons, we further examined the answers to the qualitative questions at the end of the survey. We did find that 31\% of the participants in the TSESE condition confirmed that they built their trust in the model predictions based on the source of information and evidence. This showed the effectiveness of the source of information and evidence in helping build trust, which was supported by previous research that a primary evaluation of verifiable claims was by checking the source of the information and evidence \cite{wineburg2019lateral}. However, this percentage was not big enough and it was unclear whether the addition of source and evidence could influence their trust and willingness to share information significantly on top of the model explanation augmented by SHAP. On the other hand, 16\% of the participants stated that they trusted the model predictions when they had prior knowledge about the claims (e.g, \emph{“I trust/distrust the prediction if I have previous knowledge/read about it before”}), which indicated that the pre-exposure to COVID-19 claims in the experiment could potentially mitigate the influence of extra source and evidence of the information. In order to understand the insignificance between the TSE and TSESE conditions, we further ran a one-way ANOVA (Note we first ran a 3 (explanation, i.e., T, TSE, TSESE) by 2 (claim nature, i.e., true and false) two-way ANOVA and found no main effects for claim nature or interaction effects. Then we ran a two one-way ANOVA models) to compare the three conditions for the false claims and the true claims separately. For the false claims, the main effects of the three conditions on trust ($\textit{F}(2, 241) = 7.984, \textit{p} = .000$) and willingness to share ($\textit{F}(2, 241) = 11.918, \textit{p} = .000$) were significant. As for the true claims, the main effect of the three conditions on trust ($\textit{F}(2, 241) = 2.158, \textit{p} = .118$) was not significant, but it was significant on the willingness to share ($\textit{F}(2, 241) = 6.208, \textit{p} = .002$). This indicated that model explanation help improve trust for the false claims more than the true claims, which might be explained by the fact that 11\% of the participants trusted the claims based on what was true or untrue (e.g, \emph{“I trust/distrust it based on what I know to be true or untrue”}). In this situation, there was no need for them to further check the source or evidence for true claims. Furthermore, since we only tested 10 randomly selected claims in our dataset, one should be cautious to interpret the insignificant results between the TSE and TSESE conditions. As a summary, the fact that we did not find significant differences between the TSE and TSESE conditions for trust did not necessarily indicate that sources and evidence information is not useful in helping build trust in claims associated with COVID-19. More studies are needed to further understand when sources and evidence are not needed and when they are needed in building trust in specific claims about COVID-19.

As for willingness, the majority of the participants (40\%) were willing to share only true information  (e.g, \emph{“If the information is almost guaranteed to be true, I would probably post”}). We had half of false claims in the experiment and this could potentially reduce the effects of extra sources and evidence on their willingness to share information. In addition, there were no significant differences between the TSE and TSESE conditions on willingness to share for either false claims or true claims, separately, despite the significant main effects among the three conditions. This was probably explained by the fact that 21\% of the participants in the TSESE condition were not willing to share any kind of information on social media (e.g, \emph{“I do not share any sort of information like this with my friends or family”}). Therefore, a calibration process about their tendency to share information on social media might be included in future studies to better examine the effects of extra sources and evidence on participants' willingness to share. Another possible reason could be associated with the specific content of claims about COVID-19 in the experiment and we should include more claims in future studies to further examine whether there is any difference between the TSE and TSESE conditions.

\subsection{Implications}
The risks associated with false and misleading information about COVID-19 are especially high with the rapid changing situation of the pandemic. Examples include 1) misinformation and false claims about different methods of prevention, treatment, testing, diagnosis, and miracle cures of the disease, and 2) false claims of conspiracy theories about its origins, bioweapons, and population control schemes. Thus, it is extremely important to provide a trustworthy model for the general public to verify whether such claims are true or not.  We made use of the state-of-the-art NLP machine learning models with explanations to improve both accuracy and trustworthiness of the application. As machine learning models are impacting our everyday lives, it is crucial not only to improve their performance but also to develop a better understanding of how they work. In addition, we investigated participants’ trust in the model predictions and their willingness to share the model predictions under three conditions. As our study showed, improvement in trust can be achieved through explanations offered by SHAP. To convince the public that the given information is trustworthy, we need to provide explanations of how the model made the prediction and potentially the source and evidence of the information as well. Although no significant difference was found between the TSE and TSESE conditions, further studies should be investigated to see if sources and extra evidence actually help improve trust and willingness to share information.   

\section{Conclusion and Future Work}
We built a trustworthy prediction model to debunk false claims of COVID-19 by capitalizing DistilBERT and SHAP. Our results have demonstrated the effectiveness of the proposed method and provided good implications in detecting misinformation about COVID-19 and improving public trust. Among the three conditions, participants were significantly more likely to trust and share information related to COVID-19 in the TSE and TSESE conditions than in the T condition. 

One of the limitations in building such a machine learning model is to potentially verify a large number of claims about COVID-19. Our model is built on a small dataset collected by April 2020 and the COVID-19 Fake News Detection dataset \cite{patwa2020fighting}. Thus, it might be limited to detect new misinformation related to COVID-19. In order to maintain and help improve the trustworthiness of the proposed model, it is imperative to include more data as the pandemic unfolds over time, such as the COVID-19 Healthcare Misinformation Dataset \cite{cui2020coaid}. In addition, although BERT aims to learn contextualized representation across a wide range of NLP tasks, it is still challenging to leverage BERT (i.e., it has almost no understanding of COVID-19) without domain knowledge about COVID-19. This is mainly due to the fact that there is a limited labeled number of claims about COVID-19 to fine-tune BERT to ensure full task-awareness of the system. Thus, in the future we plan to increase the domain task awareness with an unsupervised training method by making use of the COVID-19 Open Research Dataset (CORD-19) and strengthen the end task awareness using supervised fine-tuning by labeling and augmenting the claims. 
In this research, we recruited participants using AMT. Therefore, the selected sample may not be  well-representative of the population. In addition, we were not able to calibrate participants’ political and ideological biases related to COVID-19 claims, which could potentially have a significant effect on their belief and/or disbelief in such claims, although we tried to minimize such an effect through randomly assigning participants into three conditions. Future studies should include extra survey questions in order to calibrate such biases. Managing the quality of the survey data from AMT was also challenging. We removed the invalid participants by examining their responses on the three designed attention questions. However, the quality could also be affected by the compensation rate. Another limitation of this study was the limited knowledge about the participants’ demographic information, which can also influence the results in this study. Further investigation should include demographic factors. In addition, interpreting the explanations provided by SHAP can be challenging for the first time. Even though we provided a training section at the beginning of the survey, some participants found it confusing to make predictions based on individual words. In the future, more intuitive explanations should be explored to better improve trust.

% biography section

\bibliography{mybibfile}

\begin{thebibliography}{42}
\expandafter\ifx\csname natexlab\endcsname\relax\def\natexlab#1{#1}\fi
\providecommand{\url}[1]{\texttt{#1}}
\providecommand{\href}[2]{#2}
\providecommand{\path}[1]{#1}
\providecommand{\DOIprefix}{doi:}
\providecommand{\ArXivprefix}{arXiv:}
\providecommand{\URLprefix}{URL: }
\providecommand{\Pubmedprefix}{pmid:}
\providecommand{\doi}[1]{\href{http://dx.doi.org/#1}{\path{#1}}}
\providecommand{\Pubmed}[1]{\href{pmid:#1}{\path{#1}}}
\providecommand{\bibinfo}[2]{#2}
\ifx\xfnm\relax \def\xfnm[#1]{\unskip,\space#1}\fi
%Type = Misc
\bibitem[{{CDC}(8 07)}]{cdc2020}
\bibinfo{author}{{CDC}}, \bibinfo{title}{Coronavirus disease 2019
  ({COVID}-19)}, \bibinfo{year}{cited on 2020-08-07}. \URLprefix
  \url{https://www.cdc.gov/coronavirus/2019-ncov/index.html}.
%Type = Misc
\bibitem[{Kassam(5 31)}]{kassam_disinformation_2020}
\bibinfo{author}{N.~Kassam}, \bibinfo{title}{Disinformation and coronavirus:
  The dilution of information on the internet is currently posing a risk to
  global health and safety.}, \bibinfo{year}{cited on 2020-05-31}. \URLprefix
  \url{https://www.lowyinstitute.org/the-interpreter/disinformation-and-coronavirus}.
%Type = Article
\bibitem[{Beaunoyer et~al.(2020)Beaunoyer, Dup{\'e}r{\'e}, and
  Guitton}]{beaunoyer2020covid}
\bibinfo{author}{E.~Beaunoyer}, \bibinfo{author}{S.~Dup{\'e}r{\'e}},
  \bibinfo{author}{M.~J. Guitton},
\newblock \bibinfo{title}{Covid-19 and digital inequalities: Reciprocal impacts
  and mitigation strategies},
\newblock \bibinfo{journal}{Computers in Human Behavior}
  (\bibinfo{year}{2020}) \bibinfo{pages}{106424}.
%Type = Article
\bibitem[{Ozbay and Alatas(2020)}]{ozbay2020fake}
\bibinfo{author}{F.~A. Ozbay}, \bibinfo{author}{B.~Alatas},
\newblock \bibinfo{title}{Fake news detection within online social media using
  supervised artificial intelligence algorithms},
\newblock \bibinfo{journal}{Physica A: Statistical Mechanics and its
  Applications} \bibinfo{volume}{540} (\bibinfo{year}{2020})
  \bibinfo{pages}{123174}.
%Type = Article
\bibitem[{Aggarwal et~al.(2020)Aggarwal, Chauhan, Kumar, Mittal, and
  Verma}]{aggarwal2020classification}
\bibinfo{author}{A.~Aggarwal}, \bibinfo{author}{A.~Chauhan},
  \bibinfo{author}{D.~Kumar}, \bibinfo{author}{M.~Mittal},
  \bibinfo{author}{S.~Verma},
\newblock \bibinfo{title}{Classification of fake news by fine-tuning deep
  bidirectional transformers based language model},
\newblock \bibinfo{journal}{EAI Endorsed Transactions on Scalable Information
  Systems Online First; EAI: Ghent, Belgium}  (\bibinfo{year}{2020}).
%Type = Misc
\bibitem[{Sumbaly et~al.(6 21)Sumbaly, Mahalia, Shah, Khatkevich, Luo, Strauss,
  Szilvasy, Puri, Manadhata, Graham, Douze, Yalniz, and
  Jehou}]{sumbaly_using_nodate}
\bibinfo{author}{R.~Sumbaly}, \bibinfo{author}{M.~Mahalia},
  \bibinfo{author}{H.~Shah}, \bibinfo{author}{T.~Khatkevich},
  \bibinfo{author}{E.~Luo}, \bibinfo{author}{E.~Strauss},
  \bibinfo{author}{G.~Szilvasy}, \bibinfo{author}{M.~Puri},
  \bibinfo{author}{P.~Manadhata}, \bibinfo{author}{B.~Graham},
  \bibinfo{author}{M.~Douze}, \bibinfo{author}{Z.~Yalniz},
  \bibinfo{author}{H.~Jehou}, \bibinfo{title}{Using {AI} to detect {COVID}-19
  misinformation and exploitative content}, \bibinfo{year}{cited on
  2020-06-21}. \URLprefix
  \url{https://ai.facebook.com/blog/using-ai-to-detect-covid-19-misinformation-and-exploitative-content/}.
%Type = Inproceedings
\bibitem[{Devlin et~al.(2019)Devlin, Chang, Lee, and
  Toutanova}]{devlin2018bert}
\bibinfo{author}{J.~Devlin}, \bibinfo{author}{M.-W. Chang},
  \bibinfo{author}{K.~Lee}, \bibinfo{author}{K.~Toutanova},
\newblock \bibinfo{title}{Bert: Pre-training of deep bidirectional transformers
  for language understanding},
\newblock in: \bibinfo{booktitle}{Proceedings of the 2019 Conference of the
  North American Chapter of the Association for Computational Linguistics:
  Human Language Technologies, Volume 1 (Long and Short Papers)},
  \bibinfo{year}{2019}, pp. \bibinfo{pages}{4171--4186}.
%Type = Article
\bibitem[{Doshi-Velez and Kim(2017)}]{doshi2017towards}
\bibinfo{author}{F.~Doshi-Velez}, \bibinfo{author}{B.~Kim},
\newblock \bibinfo{title}{Towards a rigorous science of interpretable machine
  learning},
\newblock \bibinfo{journal}{arXiv preprint arXiv:1702.08608}
  (\bibinfo{year}{2017}).
%Type = Article
\bibitem[{Zhou et~al.(2020)Zhou, Chen, and Lei}]{zhou2020not}
\bibinfo{author}{F.~Zhou}, \bibinfo{author}{T.~Chen}, \bibinfo{author}{B.~Lei},
\newblock \bibinfo{title}{Do not forget interaction: Predicting fatality of
  covid-19 patients using logistic regression},
\newblock \bibinfo{journal}{arXiv preprint arXiv:2006.16942}
  (\bibinfo{year}{2020}).
%Type = Inproceedings
\bibitem[{Gilpin et~al.(2018)Gilpin, Bau, Yuan, Bajwa, Specter, and
  Kagal}]{gilpin2018explaining}
\bibinfo{author}{L.~H. Gilpin}, \bibinfo{author}{D.~Bau},
  \bibinfo{author}{B.~Z. Yuan}, \bibinfo{author}{A.~Bajwa},
  \bibinfo{author}{M.~Specter}, \bibinfo{author}{L.~Kagal},
\newblock \bibinfo{title}{Explaining explanations: An overview of
  interpretability of machine learning},
\newblock in: \bibinfo{booktitle}{2018 IEEE 5th International Conference on
  data science and advanced analytics (DSAA)}, \bibinfo{organization}{IEEE},
  \bibinfo{year}{2018}, pp. \bibinfo{pages}{80--89}.
%Type = Article
\bibitem[{Shapley(1953)}]{shapley1953value}
\bibinfo{author}{L.~S. Shapley},
\newblock \bibinfo{title}{A value for n-person games},
\newblock \bibinfo{journal}{Contributions to the Theory of Games}
  \bibinfo{volume}{2} (\bibinfo{year}{1953}) \bibinfo{pages}{307--317}.
%Type = Inproceedings
\bibitem[{Lundberg and Lee(2017)}]{lundberg2017unified}
\bibinfo{author}{S.~M. Lundberg}, \bibinfo{author}{S.-I. Lee},
\newblock \bibinfo{title}{A unified approach to interpreting model
  predictions},
\newblock in: \bibinfo{booktitle}{Advances in neural information processing
  systems}, \bibinfo{year}{2017}, pp. \bibinfo{pages}{4765--4774}.
%Type = Article
\bibitem[{Shu et~al.(2017)Shu, Sliva, Wang, Tang, and Liu}]{shu2017fake}
\bibinfo{author}{K.~Shu}, \bibinfo{author}{A.~Sliva},
  \bibinfo{author}{S.~Wang}, \bibinfo{author}{J.~Tang},
  \bibinfo{author}{H.~Liu},
\newblock \bibinfo{title}{Fake news detection on social media: A data mining
  perspective},
\newblock \bibinfo{journal}{ACM SIGKDD explorations newsletter}
  \bibinfo{volume}{19} (\bibinfo{year}{2017}) \bibinfo{pages}{22--36}.
%Type = Inproceedings
\bibitem[{Gilda(2017)}]{gilda2017evaluating}
\bibinfo{author}{S.~Gilda},
\newblock \bibinfo{title}{Evaluating machine learning algorithms for fake news
  detection},
\newblock in: \bibinfo{booktitle}{2017 IEEE 15th Student Conference on Research
  and Development (SCOReD)}, \bibinfo{organization}{IEEE},
  \bibinfo{year}{2017}, pp. \bibinfo{pages}{110--115}.
%Type = Inproceedings
\bibitem[{Wang(2017)}]{wang2017liar}
\bibinfo{author}{W.~Y. Wang},
\newblock \bibinfo{title}{“liar, liar pants on fire”: A new benchmark
  dataset for fake news detection},
\newblock in: \bibinfo{booktitle}{Proceedings of the 55th Annual Meeting of the
  Association for Computational Linguistics (Volume 2: Short Papers)},
  \bibinfo{year}{2017}, pp. \bibinfo{pages}{422--426}.
%Type = Inproceedings
\bibitem[{Ma et~al.(2016)Ma, Gao, Mitra, Kwon, Jansen, Wong, and
  Cha}]{ma2016detecting}
\bibinfo{author}{J.~Ma}, \bibinfo{author}{W.~Gao}, \bibinfo{author}{P.~Mitra},
  \bibinfo{author}{S.~Kwon}, \bibinfo{author}{B.~J. Jansen},
  \bibinfo{author}{K.~F. Wong}, \bibinfo{author}{M.~Cha},
\newblock \bibinfo{title}{Detecting rumors from microblogs with recurrent
  neural networks},
\newblock in: \bibinfo{booktitle}{IJCAI International Joint Conference on
  Artificial Intelligence}, volume \bibinfo{volume}{2016},
  \bibinfo{year}{2016}, pp. \bibinfo{pages}{3818--3824}.
%Type = Inproceedings
\bibitem[{Ruchansky et~al.(2017)Ruchansky, Seo, and Liu}]{ruchansky2017csi}
\bibinfo{author}{N.~Ruchansky}, \bibinfo{author}{S.~Seo},
  \bibinfo{author}{Y.~Liu},
\newblock \bibinfo{title}{Csi: A hybrid deep model for fake news detection},
\newblock in: \bibinfo{booktitle}{Proceedings of the 2017 ACM on Conference on
  Information and Knowledge Management}, \bibinfo{year}{2017}, pp.
  \bibinfo{pages}{797--806}.
%Type = Article
\bibitem[{Yang et~al.(2018)Yang, Zheng, Zhang, Cui, Li, and Yu}]{yang2018ti}
\bibinfo{author}{Y.~Yang}, \bibinfo{author}{L.~Zheng},
  \bibinfo{author}{J.~Zhang}, \bibinfo{author}{Q.~Cui},
  \bibinfo{author}{Z.~Li}, \bibinfo{author}{P.~S. Yu},
\newblock \bibinfo{title}{{TI-CNN:} convolutional neural networks for fake news
  detection},
\newblock \bibinfo{journal}{CoRR} \bibinfo{volume}{abs/1806.00749}
  (\bibinfo{year}{2018}).
%Type = Article
\bibitem[{Bahad et~al.(2019)Bahad, Saxena, and Kamal}]{bahad2019fake}
\bibinfo{author}{P.~Bahad}, \bibinfo{author}{P.~Saxena},
  \bibinfo{author}{R.~Kamal},
\newblock \bibinfo{title}{Fake news detection using bi-directional
  lstm-recurrent neural network},
\newblock \bibinfo{journal}{Procedia Computer Science} \bibinfo{volume}{165}
  (\bibinfo{year}{2019}) \bibinfo{pages}{74--82}.
%Type = Inproceedings
\bibitem[{Ma et~al.(2019)Ma, Gao, and Wong}]{ma2019detect}
\bibinfo{author}{J.~Ma}, \bibinfo{author}{W.~Gao}, \bibinfo{author}{K.-F.
  Wong},
\newblock \bibinfo{title}{Detect rumors on twitter by promoting information
  campaigns with generative adversarial learning},
\newblock in: \bibinfo{booktitle}{The World Wide Web Conference},
  \bibinfo{year}{2019}, pp. \bibinfo{pages}{3049--3055}.
%Type = Inproceedings
\bibitem[{P{\'e}rez-Rosas et~al.(2018)P{\'e}rez-Rosas, Kleinberg, Lefevre, and
  Mihalcea}]{perez2018automatic}
\bibinfo{author}{V.~P{\'e}rez-Rosas}, \bibinfo{author}{B.~Kleinberg},
  \bibinfo{author}{A.~Lefevre}, \bibinfo{author}{R.~Mihalcea},
\newblock \bibinfo{title}{Automatic detection of fake news},
\newblock in: \bibinfo{booktitle}{Proceedings of the 27th International
  Conference on Computational Linguistics}, \bibinfo{year}{2018}, pp.
  \bibinfo{pages}{3391--3401}.
%Type = Article
\bibitem[{Zhou and Zafarani(2020)}]{zhou2020survey}
\bibinfo{author}{X.~Zhou}, \bibinfo{author}{R.~Zafarani},
\newblock \bibinfo{title}{A survey of fake news: Fundamental theories,
  detection methods, and opportunities},
\newblock \bibinfo{journal}{ACM Computing Surveys (CSUR)} \bibinfo{volume}{53}
  (\bibinfo{year}{2020}) \bibinfo{pages}{1--40}.
%Type = Inproceedings
\bibitem[{Guacho et~al.(2018)Guacho, Abdali, and Papalexakis}]{guacho2018semi}
\bibinfo{author}{G.~B. Guacho}, \bibinfo{author}{S.~Abdali},
  \bibinfo{author}{E.~E. Papalexakis},
\newblock \bibinfo{title}{Semi-supervised content-based fake news detection
  using tensor embeddings and label propagation},
\newblock in: \bibinfo{booktitle}{Proc. SoCal NLP Symposium},
  \bibinfo{year}{2018}.
%Type = Inproceedings
\bibitem[{Benamira et~al.(2019)Benamira, Devillers, Lesot, Ray, Saadi, and
  Malliaros}]{benamira2019semi}
\bibinfo{author}{A.~Benamira}, \bibinfo{author}{B.~Devillers},
  \bibinfo{author}{E.~Lesot}, \bibinfo{author}{A.~K. Ray},
  \bibinfo{author}{M.~Saadi}, \bibinfo{author}{F.~D. Malliaros},
\newblock \bibinfo{title}{Semi-supervised learning and graph neural networks
  for fake news detection},
\newblock in: \bibinfo{booktitle}{2019 IEEE/ACM International Conference on
  Advances in Social Networks Analysis and Mining (ASONAM)},
  \bibinfo{organization}{IEEE}, \bibinfo{year}{2019}, pp.
  \bibinfo{pages}{568--569}.
%Type = Inproceedings
\bibitem[{Shu et~al.(2019)Shu, Wang, and Liu}]{shu2019beyond}
\bibinfo{author}{K.~Shu}, \bibinfo{author}{S.~Wang}, \bibinfo{author}{H.~Liu},
\newblock \bibinfo{title}{Beyond news contents: The role of social context for
  fake news detection},
\newblock in: \bibinfo{booktitle}{Proceedings of the twelfth ACM international
  conference on web search and data mining}, \bibinfo{year}{2019}, pp.
  \bibinfo{pages}{312--320}.
%Type = Article
\bibitem[{Dong et~al.(2020)Dong, Victor, and Qian}]{dong2020two}
\bibinfo{author}{X.~Dong}, \bibinfo{author}{U.~Victor},
  \bibinfo{author}{L.~Qian},
\newblock \bibinfo{title}{Two-path deep semi-supervised learning for timely
  fake news detection},
\newblock \bibinfo{journal}{arXiv preprint arXiv:2002.00763}
  (\bibinfo{year}{2020}).
%Type = Article
\bibitem[{Jwa et~al.(2019)Jwa, Oh, Park, Kang, and Lim}]{jwa2019exbake}
\bibinfo{author}{H.~Jwa}, \bibinfo{author}{D.~Oh}, \bibinfo{author}{K.~Park},
  \bibinfo{author}{J.~M. Kang}, \bibinfo{author}{H.~Lim},
\newblock \bibinfo{title}{exbake: Automatic fake news detection model based on
  bidirectional encoder representations from transformers (bert)},
\newblock \bibinfo{journal}{Applied Sciences} \bibinfo{volume}{9}
  (\bibinfo{year}{2019}) \bibinfo{pages}{4062}.
%Type = Article
\bibitem[{Sanh et~al.(2019)Sanh, Debut, Chaumond, and
  Wolf}]{sanh2019distilbert}
\bibinfo{author}{V.~Sanh}, \bibinfo{author}{L.~Debut},
  \bibinfo{author}{J.~Chaumond}, \bibinfo{author}{T.~Wolf},
\newblock \bibinfo{title}{Distilbert, a distilled version of bert: smaller,
  faster, cheaper and lighter},
\newblock \bibinfo{journal}{arXiv preprint arXiv:1910.01108}
  (\bibinfo{year}{2019}).
%Type = Article
\bibitem[{Xie et~al.(2019)Xie, Dai, Hovy, Luong, and Le}]{xie2019unsupervised}
\bibinfo{author}{Q.~Xie}, \bibinfo{author}{Z.~Dai}, \bibinfo{author}{E.~Hovy},
  \bibinfo{author}{M.-T. Luong}, \bibinfo{author}{Q.~V. Le},
\newblock \bibinfo{title}{Unsupervised data augmentation for consistency
  training},
\newblock \bibinfo{journal}{arXiv preprint arXiv:1904.12848}
  (\bibinfo{year}{2019}).
%Type = Inproceedings
\bibitem[{Lai and Tan(2019)}]{laihuman2019}
\bibinfo{author}{V.~Lai}, \bibinfo{author}{C.~Tan},
\newblock \bibinfo{title}{On human predictions with explanations and
  predictions of machine learning models: A case study on deception detection},
\newblock in: \bibinfo{booktitle}{Proceedings of the Conference on Fairness,
  Accountability, and Transparency}, \bibinfo{year}{2019}, pp.
  \bibinfo{pages}{29--38}.
%Type = Article
\bibitem[{Rudin(2019)}]{rudin2019stop}
\bibinfo{author}{C.~Rudin},
\newblock \bibinfo{title}{Stop explaining black box machine learning models for
  high stakes decisions and use interpretable models instead},
\newblock \bibinfo{journal}{Nature Machine Intelligence} \bibinfo{volume}{1}
  (\bibinfo{year}{2019}) \bibinfo{pages}{206--215}.
%Type = Inproceedings
\bibitem[{Kim et~al.(2016)Kim, Khanna, and Koyejo}]{kim2016examples}
\bibinfo{author}{B.~Kim}, \bibinfo{author}{R.~Khanna}, \bibinfo{author}{O.~O.
  Koyejo},
\newblock \bibinfo{title}{Examples are not enough, learn to criticize!
  criticism for interpretability},
\newblock in: \bibinfo{booktitle}{Advances in neural information processing
  systems}, \bibinfo{year}{2016}, pp. \bibinfo{pages}{2280--2288}.
%Type = Inproceedings
\bibitem[{Shu et~al.(2019)Shu, Cui, Wang, Lee, and Liu}]{shu2019defend}
\bibinfo{author}{K.~Shu}, \bibinfo{author}{L.~Cui}, \bibinfo{author}{S.~Wang},
  \bibinfo{author}{D.~Lee}, \bibinfo{author}{H.~Liu},
\newblock \bibinfo{title}{defend: A system for explainable fake news
  detection},
\newblock in: \bibinfo{booktitle}{Proceedings of the 28th ACM International
  Conference on Information and Knowledge Management}, \bibinfo{year}{2019},
  pp. \bibinfo{pages}{2961--2964}.
%Type = Book
\bibitem[{Molnar(2020)}]{molnar_chapter_nodate}
\bibinfo{author}{C.~Molnar}, \bibinfo{title}{Interpretable Machine Learning A
  Guide for Making Black Box Models Explainable}, \bibinfo{year}{2020}.
  \URLprefix
  \url{https://christophm.github.io/interpretable-ml-book/example-based.html}.
%Type = Inproceedings
\bibitem[{Reis et~al.(2019)Reis, Correia, Murai, Veloso, and
  Benevenuto}]{reis2019explainable}
\bibinfo{author}{J.~C. Reis}, \bibinfo{author}{A.~Correia},
  \bibinfo{author}{F.~Murai}, \bibinfo{author}{A.~Veloso},
  \bibinfo{author}{F.~Benevenuto},
\newblock \bibinfo{title}{Explainable machine learning for fake news
  detection},
\newblock in: \bibinfo{booktitle}{Proceedings of the 10th ACM conference on web
  science}, \bibinfo{year}{2019}, pp. \bibinfo{pages}{17--26}.
%Type = Article
\bibitem[{Patwa et~al.(2020)Patwa, Sharma, PYKL, Guptha, Kumari, Akhtar, Ekbal,
  Das, and Chakraborty}]{patwa2020fighting}
\bibinfo{author}{P.~Patwa}, \bibinfo{author}{S.~Sharma},
  \bibinfo{author}{S.~PYKL}, \bibinfo{author}{V.~Guptha},
  \bibinfo{author}{G.~Kumari}, \bibinfo{author}{M.~S. Akhtar},
  \bibinfo{author}{A.~Ekbal}, \bibinfo{author}{A.~Das},
  \bibinfo{author}{T.~Chakraborty},
\newblock \bibinfo{title}{Fighting an infodemic: Covid-19 fake news dataset},
\newblock \bibinfo{journal}{arXiv preprint arXiv:2011.03327}
  (\bibinfo{year}{2020}).
%Type = Article
\bibitem[{Wineburg and McGrew(2019)}]{wineburg2019lateral}
\bibinfo{author}{S.~Wineburg}, \bibinfo{author}{S.~McGrew},
\newblock \bibinfo{title}{Lateral reading and the nature of expertise: Reading
  less and learning more when evaluating digital information.},
\newblock \bibinfo{journal}{Teachers College Record} \bibinfo{volume}{121}
  (\bibinfo{year}{2019}) \bibinfo{pages}{n11}.
%Type = Article
\bibitem[{Lundberg et~al.(2018)Lundberg, Erion, and
  Lee}]{lundberg2018consistent}
\bibinfo{author}{S.~M. Lundberg}, \bibinfo{author}{G.~G. Erion},
  \bibinfo{author}{S.-I. Lee},
\newblock \bibinfo{title}{Consistent individualized feature attribution for
  tree ensembles},
\newblock \bibinfo{journal}{arXiv preprint arXiv:1802.03888}
  (\bibinfo{year}{2018}).
%Type = Inproceedings
\bibitem[{Kovalerchuk and Neuhaus(2018)}]{kovalerchuk2018toward}
\bibinfo{author}{B.~Kovalerchuk}, \bibinfo{author}{N.~Neuhaus},
\newblock \bibinfo{title}{Toward efficient automation of interpretable machine
  learning},
\newblock in: \bibinfo{booktitle}{2018 IEEE International Conference on Big
  Data (Big Data)}, \bibinfo{organization}{IEEE}, \bibinfo{year}{2018}, pp.
  \bibinfo{pages}{4940--4947}.
%Type = Article
\bibitem[{Ayoub et~al.(2021)Ayoub, Yang, and Zhou}]{ayoub2021modeling}
\bibinfo{author}{J.~Ayoub}, \bibinfo{author}{X.~J. Yang},
  \bibinfo{author}{F.~Zhou},
\newblock \bibinfo{title}{Modeling dispositional and initial learned trust in
  automated vehicles with predictability and explainability},
\newblock \bibinfo{journal}{Transportation Research Part F: Traffic Psychology
  and Behaviour} \bibinfo{volume}{77} (\bibinfo{year}{2021})
  \bibinfo{pages}{102--116}.
%Type = Article
\bibitem[{Mosleh et~al.(2020)Mosleh, Pennycook, and Rand}]{mosleh2020self}
\bibinfo{author}{M.~Mosleh}, \bibinfo{author}{G.~Pennycook},
  \bibinfo{author}{D.~G. Rand},
\newblock \bibinfo{title}{Self-reported willingness to share political news
  articles in online surveys correlates with actual sharing on twitter},
\newblock \bibinfo{journal}{Plos one} \bibinfo{volume}{15}
  (\bibinfo{year}{2020}) \bibinfo{pages}{e0228882}.
%Type = Article
\bibitem[{Cui and Lee(2020)}]{cui2020coaid}
\bibinfo{author}{L.~Cui}, \bibinfo{author}{D.~Lee},
\newblock \bibinfo{title}{Coaid: Covid-19 healthcare misinformation dataset},
\newblock \bibinfo{journal}{arXiv preprint arXiv:2006.00885}
  (\bibinfo{year}{2020}).

\end{thebibliography}

\end{document}